\documentclass[a4paper]{article}

\usepackage[utf8]{inputenc}
\usepackage[T1]{fontenc}
\usepackage{fouriernc}
\usepackage[english]{babel}
\usepackage{algorithm,algorithmic,array,setspace}
\usepackage{amsthm,amssymb,amsfonts,amsmath,color,mathrsfs,graphicx,float,url,breakurl,bbm,caption,subfig,placeins,xspace,eucal}
\usepackage[authoryear,round]{natbib}

\usepackage[pagebackref = true]{hyperref}
\hypersetup{
  colorlinks = true,
  urlcolor = blue, 
  linkcolor = blue,
  citecolor = blue,
  pdftitle = {Simpler PAC-Bayesian Bounds for Hostile Data - \today},
  pdfauthor = {Pierre Alquier and Benjamin Guedj},
  pdfsubject = {}
}

\newtheorem{thm}{Theorem}
\newtheorem{assum}{Assumption}
\newtheorem{dfn}{Definition}

\newtheorem{coro}{Corollary}
\newtheorem{prop}{Proposition}
\newtheorem{rmk}{Remark}
\newtheorem{exm}{Example}

\newtheoremstyle{exampstyle}
  {\topsep 3pt} % Space above
  {\topsep 3pt} % Space below
  {\itshape} % Body font
  {} % Indent amount
  {\bfseries} % Theorem head font
  {.} % Punctuation after theorem head
  {.5em} % Space after theorem head
  {\thmname{#1}\thmnumber{#2}} % Theorem head spec (can be left empty, meaning `normal')
\theoremstyle{exampstyle}

\def\Cnospace~{C{}}

\addto\extrasenglish{%

}

\allowdisplaybreaks
\graphicspath{{fig/}}

\title{Simpler PAC-Bayesian Bounds for Hostile Data}
\author{Pierre Alquier\footnote{CREST, ENSAE, Université Paris Saclay, \href{mailto:pierre.alquier@ensae.fr}{pierre.alquier@ensae.fr}. This author
gratefully acknowledges financial support from the research programme {\it New Challenges for New Data} from LCL and GENES, hosted by the {\it Fondation du Risque}, from Labex ECODEC (ANR-11-LABEX-0047) and from Labex CEMPI  (ANR-11-LABX-0007-01).}\, \& Benjamin Guedj\footnote{Modal project-team, Inria, \href{mailto:benjamin.guedj@inria.fr}{benjamin.guedj@inria.fr}.}}
\date{\today}

\begin{document}
\maketitle

\begin{abstract}
\noindent PAC-Bayesian learning bounds are of the utmost interest to the learning community. Their role is to connect
the generalization ability of an aggregation distribution $\rho$ to its
empirical risk and to its
Kullback-Leibler divergence with respect to some prior distribution $\pi$.
Unfortunately, most of the available bounds typically rely on heavy assumptions such as boundedness and independence
of the observations. This paper aims at relaxing these constraints and provides PAC-Bayesian learning bounds that
hold for dependent, heavy-tailed observations (hereafter referred to as \emph{hostile data}). In these bounds the
Kullack-Leibler divergence is replaced with a general version of
Csisz\'ar's $f$-divergence. We prove a general PAC-Bayesian bound, and show how to use it in various hostile settings.
\end{abstract}

%\tableofcontents

\section{Introduction}

Learning theory can be traced back to the late 60s and has attracted a great attention since. We refer to the monographs \cite{DGL1996} and \cite{Vap2000} for a survey. Most of the literature addresses the simplified case of i.i.d observations coupled with bounded loss functions. Many bounds on the excess risk holding with large probability were provided - these bounds are refered to as PAC learning bounds since~\cite{valiant1984theory}.\footnote{PAC stands for Probably Approximately Correct.}
\medskip

In the late 90s, the PAC-Bayesian approach was pioneered by \cite{STW} and \cite{McA,mcallester1999pac}. It consists of producing PAC bounds for a specific class of Bayesian-flavored estimators.
Similar to classical PAC results, most PAC-Bayesian bounds have been obtained with bounded loss functions \citep[see][for some of the most accurate results]{MR2483528}. Note that \cite{catoni2004statistical} provides bounds for unbouded loss, but still under very strong exponential moment assumptions. Different types of PAC-Bayesian bounds were proved in very various models~\cite{seeger2002pac,langford2002pac,seldin2010pac,seldin2012pac,seldin2011pac,guedj2013pac,begin2016pac,JMLR:v17:15-290,oneto2016pac} but the boundedness or exponential moment assumptions were essentially not improved in these papers.
\medskip

The relaxation of the exponential moment assumption is however a theoretical challenge, with huge practical implications: in many applications of regression, there is no reason to believe that the noise is bounded or sub-exponential. Actually, the belief that the noise is sub-exponential leads to an overconfidence in the prediction that is actually very harmful in practice, see for example the discussion in~\cite{taleb2007black} on finance. Still, thanks to the aforementionned works, the road to obtain PAC bounds for bounded observations has now become so nice and comfortable that it might refrain inclination to explore different settings.
\medskip

Regarding PAC bounds for heavy-tailed random variables, let us mention three recent approaches.
\begin{itemize}
\item Using the so-called {\it small-ball property}, Mendelson and several co-authors developed in a striking series of papers tools to study the Empirical Risk Minimizer (ERM) and penalized variants without an exponential moment assumption: we refer to their most recent works \citep{mendelsonlearning2015,lecuemendelson2016regularization}. Under a quite similar assumption, \cite{grunwald2016fast} derived PAC-Bayesian learning bounds (``empirical witness of badness'' assumption). Other assumptions were introduced in order to derive fast rates for unbounded losses, like the multiscale Bernstein assumption~\cite{NIPS2016_6104}.
\item Another idea consists in using robust loss functions. This leads to better confidence bounds than the previous approach, but at the price of replacing the ERM by a more complex estimator, usually building on PAC-Bayesian approaches \citep{audibert2011robust,catoni2012challenging,oliveira2013lower,ilaria2015,catoni2016pac}.
\item Finally, \cite{devroye2015sub}, using median-of-means, provide bounds in probability for the estimation of the mean without exponential moment assumption. It is possible to extend this technique to more general learning problems~\cite{minsker2015geometric,hsu2016loss,lugosi2016risk,guillaume2017learning,lugosi2017regularization}.
\end{itemize}
Leaving the well-marked path of bounded variables led the authors to sophisticated and technical mathematics, but in the end they obtained rates of convergence similar to the ones in bounded cases: this is highly valuable for the statistical and machine learning community.
\medskip

Regarding dependent observations, like time series or random fields, PAC and/or PAC-Bayesian bounds were provided in various settings~\citep{modha1998memory,steinwart2009fast,mohri2010stability,ralaivola2010chromatic,seldin2012pac,alquier2012model,alquier2012prediction,agarwal2013generalization,alquier2013prediction,kuznetsov2014generalization,giraud2015aggregation,zimin2015conditional,london2016stability}. However these works massively relied on concentration inequalities for or limit theorems for time series~\cite{yu1994rates,Doukhan1994,Rio2000,kontorovich2008concentration}, for which boundedness or exponential moments are crucial.
\medskip

This paper shows that a proof scheme of PAC-Bayesian bounds proposed by~\cite{begin2016pac} can be extended to a very general setting, without independence nor exponential moments assumptions. We would like to stress that this approach is not comparable to the aforementionned work, and in particular it is technically far less sophisticated. However, while it leads to sub-optimal rates in many cases, it allows to derive PAC-Bayesian bounds in settings where no PAC learning bounds were available before: for example heavy-tailed time series.
\medskip

Given the simplicity of the main result, we state it in the remainder of this section. The other sections are devoted to refinements and applications.
Let $\ell$ denote a generic loss function. The observations are denoted
$(X_1,Y_1),\dots,(X_n,Y_n)$. Note that we do not require the observations to be independent,
nor indentically distributed. We assume that a family of
predictors $(f_{\theta},\theta\in\Theta)$ is chosen. Let $\ell_i(\theta)=\ell[f_{\theta}(X_i),Y_i]$, and define the (empirical) risk as
\begin{align*}
r_n(\theta) & = \frac{1}{n}\sum_{i=1}^n \ell_i(\theta), \\
R(\theta) & = \mathbb{E}\bigl[r_n(\theta)\bigr].
\end{align*}
Based on the observations, the objective is to build procedures with a small risk $R$. While PAC bounds focus on estimators $\hat{\theta}_n$ that are obtained as functionals of the sample, the PAC-Bayesian approach studies an aggregation distribution $\hat{\rho}_n$ that depends on the sample. In this case, the objective is to choose $\hat{\rho}_n$ such that $\int R(\theta) \hat{\rho}_n({\rm d}\theta)$ is small. In order to do so, a crucial point is to choose a reference probability measure $\pi$, often referred to as the {\it prior}. In~\cite{MR2483528}, the role of $\pi$ is discussed in depth: rather than reflecting a prior knowledge on the parameter space $\Theta$, it should serve as a tool to measure the complexity of $\Theta$.
\medskip

Let us now introduce the two following key quantities.
\begin{dfn}
For any function $g$, let
$$
\mathcal{M}_{g,n} =  \int \mathbb{E}\bigl[g \left( |r_n(\theta)-R(\theta)| \right)
\bigr] \pi({\rm d}\theta).
$$
\end{dfn}

\begin{dfn}
Let $f$ be a convex function with $f(1)=0$. The $f$-divergence between two distributions $\rho$ and $\pi$ is defined by
$$ D_{f}(\rho,\pi)
   = \int f\left(\frac{{\rm d}\rho}{{\rm d}\pi}\right) {\rm d}\pi $$
when $\rho$ is absolutely continous with respect to $\pi$, and
$ D_{f}(\rho,\pi) = +\infty$
otherwise.
\end{dfn}
Csisz\'ar introduced $f$-divergences in the 60s, see 
his recent monograph~\citet[][Chapter 4]{csiszar2004information} for a survey.
\medskip

We use the following notation for recurring functions:
$\phi_p(x) = x^p$.% and $\psi_p(x) = \exp(x^p)-1$.
Consequently
$\mathcal{M}_{\phi_p,n} =  \int \mathbb{E}\left( |r_n(\theta)-R(\theta)|^{p}
\right) \pi({\rm d}\theta)$. Thus $\mathcal{M}_{\phi_p,n}$ is a moment of order $p$. As for divergences, we denote the Kullback-Leibler divergence by $\mathcal{K}(\rho,\pi)=D_{f}(\rho,\pi)$
when $f(x)=x\log(x)$, and the chi-square divergence $\chi^2(\rho,\pi)=D_{\phi_2-1}(\rho,\pi)$.

\begin{thm}
\label{theorem}
 Fix $p>1$, put $q=\frac{p}{p-1}$ and fix $\delta\in(0,1)$.
 With probability at least $1-\delta$ we have for any aggregation distribution $\rho$
 \begin{equation}
 \label{eq-thm-0}
 \left| \int R {\rm d}\rho -
 \int r_n {\rm d}\rho \right|
 \leq
 \left( \frac{\mathcal{M}_{\phi_{q},n} }{\delta}\right)^{\frac{1}{q}}
  \left(D_{\phi_p-1}(\rho,\pi) +1 \right)^{\frac{1}{p}}.
 \end{equation}
\end{thm}

The main message of \autoref{theorem} is that we can compare $\int r_n {\rm d}\rho$ (observable) to
$\int R {\rm d}\rho$ (unknown, the objective) in terms of
two quantities: the moment $\mathcal{M}_{\phi_{q},n}$ (which depends on the
distribution of the data) and the divergence
$D_{\phi_p-1}(\rho,\pi)$ (which will reveal itself as a measure of the complexity
of the set $\Theta$). The most important practical consequence is that we have, with probability at least $1-\delta$, for any probability measure $\rho$,
\begin{equation}
\label{theorem-equation}
\int R {\rm d}\rho \leq
 \int r_n {\rm d}\rho
 +
 \left( \frac{\mathcal{M}_{\phi_{q},n} }{\delta}\right)^{\frac{1}{q}}
  \left(D_{\phi_p-1}(\rho,\pi) +1 \right)^{\frac{1}{p}}.
\end{equation}
This is a strong incitement to define our aggregation distribution $\hat{\rho}_n$
as the minimizer of the right-hand side of \eqref{theorem-equation}.
The core of the paper will discuss in details
this strategy and other consequences of \autoref{theorem}.
\begin{proof}[Proof of \autoref{theorem}]
Introduce $\Delta_n(\theta):= |r_n(\theta)-R(\theta)|$.
We follow a scheme of proof introduced by~\cite{begin2016pac} in the bounded setting. We adapt the proof to the general case:
\begin{align*}
 \left| \int R {\rm d}\rho -
 \int r_n {\rm d}\rho \right|
 & \leq
 \int \Delta_{n} {\rm d}\rho
 = \int \Delta_{n} \frac{{\rm d}\rho}{{\rm d}\pi} {\rm d}\pi
 \\
 &
 \leq\left( \int \Delta_{n}^{q}{\rm d}\pi \right)^{\frac{1}{q}}
  \left( \int \left(\frac{{\rm d}\rho}{{\rm d}\pi}\right)^{p}   {\rm d}\pi \right)^{\frac{1}{p}} \text{ (H\"older ineq.)}
 \\
 &
 \leq\left( \frac{\mathbb{E} \int \Delta_{n}^{q}{\rm d}\pi}{\delta} \right)^{\frac{1}{q}}
  \left( \int \left(\frac{{\rm d}\rho}{{\rm d}\pi}\right)^{p}   {\rm d}\pi \right)^{\frac{1}{p}}  \text{ (Markov ineq., w. prob. } 1-\delta)
 \\
 &
 =\left( \frac{\mathcal{M}_{\phi_{q},n} }{\delta}\right)^{\frac{1}{q}}
  \left(D_{\phi_p-1}(\rho,\pi) +1 \right)^{\frac{1}{p}} .
\end{align*}
\end{proof}

In \autoref{section-divergence} we discuss the divergence term
$D_{\phi_p-1}(\rho,\pi)$.
In particular, we derive an explicit bound on this term
when $\rho$ is chosen in order to concentrate
around the ERM (empirical risk minimizer)
$\hat{\theta}_{{\rm ERM}}=\arg\min_{\theta\in\Theta}\ r_n(\theta)$.
This is meant to provide the reader some intuition on the order of magnitude of
the divergence term.
In \autoref{section-moments} we discuss how to control the moment
$\mathcal{M}_{\phi_{q},n}$. We derive explicit bounds in various examples:
bounded and unbounded losses, i.i.d and dependent observations. The most important result of the section is a risk bound for auto-regression with heavy-tailed time series, something new up to our knowledge. In \autoref{section-oracle} we come back to the general case. We show that it is possible to explicitely minimize the right-hand side in~\eqref{theorem-equation}. We then show that \autoref{theorem}
leads to powerful oracle inequalities in the various statistical settings
discussed above, exhibiting explicit rates of convergence.

\section{Calculation of the divergence term}
\label{section-divergence}

The aim of this section is to provide some hints on the order of magnitude of
the divergence term $D_{\phi_p-1}(\rho,\pi)$. We start with the example of a
finite parameter space $\Theta$. The following proposition results from straightforward
calculations.
\begin{prop}
Assume that ${\rm Card}(\Theta)=K<\infty$ and that $\pi$ is uniform on $\Theta$. Then
$$
D_{\phi_p-1}(\rho,\pi) +1= K^{p-1} \sum_{\theta\in\Theta} \rho(\theta)^p.
$$
\end{prop}
A special case of interest is when $\rho=\delta_{\hat{\theta}_{{\rm ERM}}}$,
the Dirac mass concentrated on the ERM.
Then
$$
D_{\phi_p-1}(\delta_{\hat{\theta}_{{\rm ERM}}},\pi)+1 = K^{p-1}.
$$
Then~\eqref{eq-thm-0} in~\autoref{theorem} yields the following result.
\begin{prop}
\label{prop-finite}
 Fix $p>1$, $q=\frac{p}{p-1}$ and $\delta\in(0,1)$.
 With probability at least $1-\delta$ we have
$$
R(\hat{\theta}_{{\rm ERM}})
\leq \inf_{\theta\in\Theta}\ \bigl\{r_n(\theta)\bigr\} +
K^{1-\frac{1}{p}}
\left( \frac{\mathcal{M}_{\phi_{q},n} }{\delta}\right)^{\frac{1}{q}}.
$$
\end{prop}
Remark that $D_{\phi_p-1}(\rho,\pi)$ seems to be related to the complexity $K$ of the parameter space $\Theta$. This intuition can be extended to an infinite parameter space, for example using the empirical complexity parameter introduced in~\cite{MR2483528}.
\begin{assum}
\label{empirical-complexity}
There exists $d>0$ such that, for any $\gamma>0$,
$$ \pi \Bigl\{ \theta\in\Theta: \bigl\{ r_n(\theta ) \bigr\} \leq \inf_{\theta'\in\Theta}\ r_n(\theta')
+ \gamma \Bigr\}
\geq \gamma^d .$$
\end{assum}
In many examples, $d$ corresponds to the ambient dimension
(see~\cite{MR2483528} for a thorough discussion).
In this case, a sensible choice for $\rho$, as suggested by Catoni,
is $\pi_{\gamma} ({\rm d}\theta) \propto \pi({\rm d}\theta) \mathbf{1}\left[r(\theta)
- r_n(\hat{\theta}_{{\rm ERM}}) \leq \gamma\right] $ for $\gamma$ small enough
(in~\autoref{section-oracle}, we derive the consequences
of \autoref{empirical-complexity} for other aggregation distributions).
We have
$$
D_{\phi_p-1}(\pi_{\gamma},\pi) +1  \leq \gamma^{-d(p-1)}
$$
and
$$ \int r_n (\theta) {\rm d}\pi_{\gamma} \leq r_n(\hat{\theta}_{{\rm ERM}}) + \gamma $$
so \autoref{theorem} leads to
 $$
 \int R {\rm d}\pi_{\gamma}
 \leq
 r_n(\hat{\theta}_{{\rm ERM}}) + \gamma
 +
 \gamma^{-\frac{d}{q}}
 \left(\frac{ \mathcal{M}_{\phi_{q},n}}{\delta} \right)^{\frac{1}{q}}.
 $$
 An explicit optimization with respect to $\gamma$ leads to the choice
 $$\gamma=\left(\frac{d}{q}
 \frac{\mathcal{M}_{\phi_{q},n}}{\delta}\right)^{\frac{1}{1+\frac{d}{q}}} $$
 and consequently to the following result.
 \begin{prop}
 Fix $p>1$, $q=\frac{p}{p-1}$ and $\delta\in(0,1)$.
 Under~\autoref{empirical-complexity},
 with probability at least $1-\delta$ we have,
\begin{equation*}
 \int R {\rm d}\pi_{\gamma}
 \leq \inf_{\theta\in\Theta}\ \Bigl\{r_n(\theta)\Bigr\}  
+
\left(  \frac{\mathcal{M}_{\phi_{q},n}}{\delta} \right)^{\frac{1}{1+\frac{d}{q}}}
\left\{ \left(\frac{d}{q}\right)^\frac{1}{1+\frac{d}{q}}
+ \left(\frac{d}{q}\right)^{\frac{-\frac{d}{q}}{1+\frac{d}{q}}}
\right\}.
\end{equation*}
\end{prop}

So the bound is in $\mathcal{O}((\mathcal{M}_{\phi_{q},n}/\delta)^{1/(1+d/q)})$. In order to understand the order of magnitude of the bound, it is now crucial
to understand the moment term $\mathcal{M}_{\phi_{q},n}$. This is the object
of the next section.

\section{Bounding the moments}
\label{section-moments}

In this section, we show how to control $\mathcal{M}_{\phi_{q},n}$ depending on the assumptions on the data.

\subsection{The i.i.d setting}

First, let us assume that the observations $(X_i, Y_i)$ are independent and identically distributed.
In general, when the observations are possibly heavy-tailed, we recommend to use \autoref{theorem} with $q\leq 2$ (which implies $p\geq 2)$.

\begin{prop}
\label{prop-moment}
Assume that
$$ s^2 = \int {\rm Var}[\ell_1(\theta)] \pi({\rm d}\theta)< +\infty $$
then
$$\mathcal{M}_{\phi_{q},n} \leq  \left(\frac{ s^2 }{n}\right)^{\frac{q}{2}}. $$
\end{prop}

As a conclusion for the case $q \leq 2 \leq p$,~\eqref{eq-thm-0}
in~\autoref{theorem} becomes:
$$
 \int R {\rm d}\rho
 \leq
 \int r_n {\rm d}\rho
 + \frac{\left(D_{\phi_{p}-1}(\rho,\pi)
 +1 \right)^{\frac{1}{p}}}{\delta^{\frac{1}{q}}}
   \sqrt{\frac{s^2}{n}}.
$$
Without further assumptions, this bound can not be improved as a function of $n$ (as can be seen in the simplest case where $\rm{card}(\Theta)= 1$, by using the CLT).

\begin{proof}[Proof of~\autoref{prop-moment}]
\begin{align*}
\mathcal{M}_{\phi_{q},n}
&
=  \int \mathbb{E}\left( |r_n(\theta)-\mathbb{E}[r_n(\theta)]|^{2\frac{q}{2}} \right) \pi({\rm d}\theta)
\\
&
\leq \left( \int \mathbb{E}\left( |r_n(\theta)-\mathbb{E}[r_n(\theta)]|^{2} \right) \pi({\rm d}\theta) \right)^{\frac{q}{2}}
\\
&
\leq \left( \int \frac{1}{n}{\rm Var}[\ell_1(\theta)] 
     \pi({\rm d}\theta) \right)^{\frac{q}{2}}
=  \left(\frac{ s^2 }{n}\right)^{\frac{q}{2}}.
\end{align*}
\end{proof}

As an example,
consider the regression setting with quadratic loss, where we use linear predictors: $X_i\in\mathbb{R}^k$, $\Theta=\mathbb{R}^k$ and $f_{\theta}(\cdot)=\left<\cdot,\theta\right>$.
Define a prior $\pi$ on $\Theta$ such that
\begin{equation}
\label{kurtosis-prior}
\tau := \int \|\theta\|^4 \pi({\rm d}\theta) < \infty
\end{equation}
and assume that
\begin{equation}
\label{kurtosis}
\kappa :=
 8[\mathbb{E}(Y_i^4) + \tau \mathbb{E}(\|X_i\|^4)] < \infty .
\end{equation}
Then
$$
\ell_i(\theta) = (Y_i - \left<\theta,X_i\right>)^2
\leq 2 \left[ Y_i^2 +  \|\theta\|^2 \|X_i\|^2 \right]
$$
and so
$$
{\rm Var}(\ell_i(\theta))
\leq \mathbb{E}(\ell_i(\theta)^2)
 \leq 8 \mathbb{E}\left[ Y_i^4 +  \|\theta\|^4 \|X_i\|^4 \right].
$$
Finally,
$$
s^2
 = \int {\rm Var}(\ell_i(\theta)) \pi({\rm d}\theta)
 \leq \kappa<+\infty.
$$

We obtain the following corollary of~\eqref{eq-thm-0} in~\autoref{theorem} with $p=q=2$.

\begin{coro}
 Fix $\delta\in(0,1)$. Assume that $\pi$ is chosen such that~\eqref{kurtosis-prior}
 holds, and assume that~\eqref{kurtosis} also holds.
 With probability at least $1-\delta$ we have for any $\rho$
 $$
 \int R {\rm d}\rho
 \leq
 \int r_n {\rm d}\rho
 + \sqrt{ \frac{\kappa[1+\chi^2(\rho,\pi)]}{n \delta}}.
 $$
\end{coro}

Note that a similar upper bound was proved in~\cite{honorio2014tight},
yet only in the case of the 0-1 loss (which is bounded). Also, note
that the assumption on the moments of order $4$ is comparable to the one in~\cite{audibert2011robust} and allow heavy-tailed distributions. Still, in our result, the dependence in $\delta$ is less good than in~\cite{audibert2011robust}. So, we end this subsection with a study of the sub-Gaussian case (wich also includes the bounded case). In this case, we can use any $q\geq 2$ in \autoref{theorem}. The larger
$q$, the better will be the dependence with respect to $\delta$.

\begin{dfn}
 A random variable $U$ is said to be sub-Gaussian with parameter $\sigma^2$ if for any $\lambda>0$,
$$
\mathbb{E}\Bigl\{\exp\bigl[ \lambda(U-\mathbb{E}(U))
 \bigr]\Bigr\}
\leq \exp\left[ \frac{\lambda^2 \sigma^2}{2} \right].
$$
\end{dfn}
\begin{prop}[Theorem 2.1 page 25 in~\cite{boucheron2013concentration}]
When $U$ is sub-Gaussian with parameter $\sigma^2$ then for any $q\geq 2$,
$$
\mathbb{E}\bigl[(U-\mathbb{E}(U))^q\bigr] \leq 2 \left(\frac{q}{2}\right)! (2\sigma^2)^{\frac{q}{2}} \leq 2 (q \sigma^2)^{\frac{q}{2}}.
$$
\end{prop}
A straighforward consequence is the following result.
\begin{prop}
\label{prop-exp}
 Assume that, for any $\theta$, $\ell_i(\theta)$ is sub-Gaussian with parameter
 $\sigma^2$ (that does not depend on $\theta$), then  $\frac{1}{n}\sum_{i=1}^n \ell_i(\theta)$ is sub-Gaussian with parameter $\sigma^2/n$ and then, for any $q\geq 2$,
$$
\mathcal{M}_{\phi_{q},n}
\leq 2 \left(\frac{q \sigma^2}{n} \right)^{\frac{q}{2}}.
$$
\end{prop}
As an illustration, consider the case of a finite parameter space, that is ${\rm card}(\Theta)=K<+\infty$. Following~\autoref{prop-finite} and~\autoref{prop-exp}, we obtain for any $q\geq 2$ and $\delta\in(0,1)$, with probability at least $1-\delta$,
$$
R(\hat{\theta}_{{\rm ERM}})
\leq \inf_{\theta\in\Theta}\ \bigl\{r_n(\theta)\bigr\} +
\sigma \sqrt{\frac{q}{n}}
\left(\frac{2K}{\delta}\right)^{\frac{1}{q}}.
$$
Optimization with respect to $q$ leads to $q=2\log(2K/\delta)$
and consequently
$$
R(\hat{\theta}_{{\rm ERM}})
\leq \inf_{\theta\in\Theta}\ \bigl\{r_n(\theta)\bigr\}+
 \sqrt{\frac{2 {\rm e}\sigma^2 \log\left(\frac{2K}{\delta}\right)}{n}}.
$$
Without any additional assumption on the loss $\ell$, the rate on the right-hand side is optimal. This is for example proven by~\cite{audibert2009fast} for the absolute loss.

\subsection{Dependent observations}

Here we propose to analyze the harder and more realistic case where the observations $(X_i,Y_i)$ are possibly dependent. It includes the autoregressive case where $X_i=Y_{i-1}$ or $X_i=(Y_{i-1},\dots,Y_{i-p})$. Note that in this setting, different notions of risks were used in the literature. The risk $R(\theta)$ considered in this paper is the same as the one used in many references given in the introduction, \cite{modha1998memory,steinwart2009fast,alquier2012model,alquier2012prediction} among others. Alternative notions of risk were proposed, for example by~\cite{zimin2015conditional}.
\medskip

We remind the following definition.
\begin{dfn}
 The $\alpha$-mixing coefficients between two $\sigma$-algebras $\mathcal{F}$ and $\mathcal{G}$ are defined by
 $$ \alpha(\mathcal{F},\mathcal{G}) = \sup_{A\in\mathcal{F},B\in\mathcal{G}}
          \Bigl|\mathbb{P}(A \cap B) - \mathbb{P}(A) \mathbb{P}(B)\bigr|. $$
\end{dfn}
 We refer the reader to~\cite{Doukhan1994} and \cite{Rio2000} (among others) for more details. We still provide a basic interpretation of this definition. First, when $\mathcal{F}$ and $\mathcal{G}$ are independent, then for all $A\in\mathcal{F}$ and $B\in\mathcal{G}$, $\mathbb{P}(A \cap B) = \mathbb{P}(A) \mathbb{P}(B)$ by definition of independence, and so $\alpha(\mathcal{F},\mathcal{G})=0$. On the other hand, when $\mathcal{F}=\mathcal{G}$, as soon as these $\sigma$-algebras contain an event $A$ with $\mathbb{P}(A)=1/2$ then $\alpha(\mathcal{F},\mathcal{G})= |\mathbb{P}(A \cap A) - \mathbb{P}(A) \mathbb{P}(A)| = |1/2-1/4|=1/4$. More generally, $\alpha(\mathcal{F},\mathcal{G})$ is a measure of the dependence of the information provided by $\mathcal{F}$ and $\mathcal{G}$, ranging from $0$ (independance) to $1/4$ (maximal dependence). We provide another interpretation in terms of covariances.
\begin{prop}[Classical, see~\cite{Doukhan1994} for a proof]
 We have
 \begin{multline*} \alpha(\mathcal{F},\mathcal{G})
  = \sup\Bigl\{{\rm Cov}(U,V), 0\leq U \leq 1, 0\leq V\leq 1,
  \\
  U \text{ is } \mathcal{F}\text{-measurable, }V \text{ is } \mathcal{G}\text{-measurable}\Bigr\}.
 \end{multline*}
\end{prop}
For short, define
$$\alpha_j=\alpha[\sigma(X_0,Y_0),\sigma(X_j,Y_j)] $$
where we remind that for any random variable $Z$, $\sigma(Z)$ is the $\sigma$-algebra generated by $Z$. The idea is that, when the future of the series is strongly dependent of the past, $\alpha_j$ will remain constant, or decay very slowly. On the other hand, when the near future is almost independent of the past, then the $\alpha_j$ decay very fast to $0$ (examples of both kind can be found in~\cite{Doukhan1994,Rio2000}). And, indeed, we will see below that when the rate of convergence of the $\alpha_j$'s to $0$ is fast enough it is possible to derive results rather similar to the ones the independent case.

\medskip

Let us first consider the bounded case.
\begin{prop}
\label{proposition-seriestemp-bornees}
Assume that $0\leq \ell\leq 1$.
Assume that $(X_i,Y_i)_{i\in\mathbb{Z}}$ is a stationary process, and that it satisfies $
\sum_{j\in\mathbb{Z}} \alpha_j < \infty$.
Then
$$  \mathcal{M}_{\phi_{2},n} \leq \frac1{n}\sum_{j\in\mathbb{Z}} \alpha_{j} .$$
\end{prop}

Examples of processes satisfying this assumption are discussed in~\cite{Doukhan1994,Rio2000}. For example, if the $(X_i,Y_i)$'s are actually a geometrically ergodic Markov chain then there exist some $c_1, c_2>0$ such that
$\alpha_j \leq c_1 {\rm e}^{-c_2 |j|}$. Thus
$$\mathcal{M}_{\phi_{2},n}\leq \frac{1}{n} \frac{2c_1}{1-{\rm e}^{-c_2}} .$$

\begin{proof}[Proof of~\autoref{proposition-seriestemp-bornees}]
We have:
\begin{align*}
\mathbb{E} \left[ \left(\frac{1}{n}\sum_{i=1}^n \ell_{i}(\theta) - \mathbb{E}[\ell_{i}(\theta)]\right)^2\right]
& = \frac{1}{n^2} \sum_{i=1}^n \sum_{j=1}^n {\rm Cov}[\ell_{i}(\theta),\ell_{j}(\theta)]
 \\
 & \leq \frac{1}{n^2}  \sum_{i=1}^n \sum_{j\in\mathbb{Z}} \alpha_{j-i}
     = \frac{\sum_{j\in\mathbb{Z}} \alpha_{j}}{n} 
\end{align*}
that does not depend on $\theta$, and so
$$ \mathcal{M}_{\phi_{2},n} = \int \mathbb{E} \left[ \left(\frac{1}{n}\sum_{i=1}^n \ell_{i}(\theta) - R(\theta)\right)^2\right]
        \pi({\rm d}\theta) \leq \frac{\sum_{j\in\mathbb{Z}} \alpha_{j}}{n} .$$
\end{proof}

\begin{rmk}
Other assumptions than $\alpha$-mixing can be used. Actually, we see from the proof that
the only requirement to get a bound on $\mathcal{M}_{\phi_{2},n}$ is to control the
covariance ${\rm Cov}[\ell_{i}(\theta),\ell_{j}(\theta)]$; $\alpha$-mixing is very stringent
as it imposes that we can control this for any function $\ell_i(\theta)$. In the case of
a Lipschitz loss, we could actually consider more general conditions like the weak dependence conditions in~\cite{dedecker2007weak,alquier2012model}.
\end{rmk}

We now turn to the unbounded case.
\begin{prop}
\label{proposition-seriestemp-unbounded}
Assume that $(X_i,Y_i)_{i\in\mathbb{Z}}$ is a stationary process. Let $r \geq 1$ and $s \geq 2$ be any numbers with $1/r+2/s=1$ and assume that
$$
\sum_{j\in\mathbb{Z}} \alpha_j^{1/r} < \infty
$$
and
$$
\int  \left\{\mathbb{E}\left[ \ell_{i}^s(\theta)\right]\right\}^{\frac{2}{s}} \pi({\rm d}\theta) < \infty.
$$
Then
$$ \mathcal{M}_{\phi_{2},n} \leq \frac{1}{n} \left(
\int  \left\{\mathbb{E}\left[ \ell_{i}^s(\theta)\right]\right\}^{\frac{2}{s}} \pi({\rm d}\theta) \right)
\left(\sum_{j\in\mathbb{Z}} \alpha_{j}^{\frac{1}{r}}\right) .$$
\end{prop}

\begin{proof}[Proof of \autoref{proposition-seriestemp-unbounded}]
The proof relies on the following property.
\begin{prop}[\cite{Doukhan1994}]
 For any random variables $U$ and $V$, resp. $\mathcal{F}$ and $\mathcal{G}$-mesurable, we have
 $$
 |{\rm Cov}(U,V)| \leq 8 \alpha^{\frac{1}{r}}(\mathcal{F},\mathcal{G}) \|U\|_s \|V\|_t
 $$
 where $1/r + 1/s + 1/t = 1$.
\end{prop}
We use this with $U=\ell_{i}(\theta) $, $V=\ell_{j}(\theta)$
and $s=t$. Then
\begin{align*}
\mathbb{E} \Biggl[ \Biggl(\frac{1}{n}\sum_{i=1}^n \ell_{i}(\theta)
  - \mathbb{E}[\ell_{i}(\theta)]\Biggr)^2\Biggr]
 & = \frac{1}{n^2} \sum_{i=1}^n \sum_{j=1}^n {\rm Cov}[\ell_{i}(\theta),\ell_{j}(\theta)]
 \\
  & \leq \frac{8}{n^2}  \sum_{i=1}^n \sum_{j\in\mathbb{Z}} \alpha_{j-i}^{\frac{1}{r}} \|\ell_{i}(\theta)\|_s \|\ell_{j}(\theta)\|_s
 \\
& \leq \frac{ 8\left\{ \mathbb{E}\left[ \ell_{i}^s(\theta)\right]\right\}^{\frac{2}{s}} \sum_{j\in\mathbb{Z}} \alpha_{j}^{\frac{1}{r}}}{n}.
\end{align*}
\end{proof}

As an example, consider auto-regression with quadratic loss, where we use linear predictors: $X_i=(1,Y_{i-1})\in\mathbb{R}^2$, $\Theta=\mathbb{R}^2$ and $f_{\theta}(\cdot)=\left<\theta,\cdot\right>$. Then
$$
|\ell_{i}(\theta)|^3 \leq 32 [Y_i^6 + 4\|\theta\|^6(1+ Y_{i-1}^6) ]
$$
and so
$$
\mathbb{E}\left(|\ell_{i}(\theta)|^3\right) \leq 32(1+4\|\theta\|^6 )\mathbb{E}\left(Y_i^6\right).
$$
Taking $s=r=3$ in \autoref{proposition-seriestemp-unbounded} leads to the following result.
\begin{coro}
\label{coro:timeseries}
 Fix $\delta\in(0,1)$. Assume that $\pi$ is chosen such that
 $$
 \int \|\theta\|^6 \pi({\rm d}\theta) < + \infty,
 $$
 $\mathbb{E}\left(Y_i^6\right)<\infty$ and $\sum_{j\in\mathbb{Z}} \alpha_j^{\frac{1}{3}} < +\infty$.
 Put
 $$ \nu = 32 \mathbb{E}\left(Y_i^6\right)^{\frac{2}{3}} \sum_{j\in\mathbb{Z}} \alpha_j^{\frac{1}{3}} \left( 1+ 4 \int\|\theta\|^6 \pi({\rm d}\theta)\right) . $$
 With probability at least $1-\delta$ we have for any $\rho$
 $$
 \int R {\rm d}\rho
 \leq
 \int r_n {\rm d}\rho
 + \sqrt{ \frac{\nu[1+\chi^2(\rho,\pi)]}{n \delta}}.
 $$
\end{coro}
This is, up to our knowledge, the first PAC(-Bayesian) bound in the case of a time series without any boundedess nor exponential moment assumption.

\section{Optimal aggregation distribution and oracle inequalities}
\label{section-oracle}

We have now gone through the way to control the different terms in our PAC-Bayesian inequality (\autoref{theorem}). We now come back to this result to derive which predictor minimizes the bound, and which statistical guarantees can be achieved by this predictor.
\medskip

 We start with a reminder of two consequences of~\autoref{theorem}:
 for $p>1$, and $q=p/(p-1)$,
 with probability at least $1-\delta$ we have for any $\rho$
 \begin{equation}
 \label{eq-thm-1}
 \int R {\rm d}\rho
 \leq
 \int r_n {\rm d}\rho
 + \left( \frac{\mathcal{M}_{\phi_{q},n} }{\delta}\right)^{\frac{1}{q}}
  \left(D_{\phi_p-1}(\rho,\pi) +1 \right)^{\frac{1}{p}}
 \end{equation}
 and
  \begin{equation}
 \label{eq-thm-2}
 \int r_n {\rm d}\rho
 \leq
 \int R {\rm d}\rho
 + \left( \frac{\mathcal{M}_{\phi_{q},n} }{\delta}\right)^{\frac{1}{q}}
  \left(D_{\phi_p-1}(\rho,\pi) +1 \right)^{\frac{1}{p}}.
 \end{equation}

In this section we focus on
the minimizer $\hat{\rho}_n$ of the right-hand side of~\eqref{eq-thm-1} , and on its statistical properties.

\begin{dfn}
 We define $\overline{r}_n=\overline{r}_n(\delta,p)$ as
$$
\overline{r}_n = \min\left\{u\in\mathbb{R},
\int
\left[u -r_n(\theta)\right]_+^{q}
\pi({\rm d}\theta) = \frac{\mathcal{M}_{\phi_{q},n} }{\delta}\right\}.
$$
Note that such a minimum always exists as the integral is a
continuous function of $u$,
is equal to $0$ when $u=0$ and $\rightarrow \infty$ when $u\rightarrow\infty$.
We then define
\begin{equation}
\label{definitionRho}
\frac{{\rm d}\hat{\rho}_n}{{\rm d}\pi}(\theta) = \frac{
\left[\overline{r}_n -r_n(\theta)\right]_+^{\frac{1}{p-1}}
}{
\int \left[\overline{r}_n -r_n\right]_+^{\frac{1}{p-1}} {\rm d}\pi
}.
\end{equation}
\end{dfn}

The following proposition states that $\hat{\rho}_n$ is actually the minimizer of the right-hand side in inequality \eqref{eq-thm-1}.

\begin{prop}
\label{prop-minimizer}
Under the assumptions of~\autoref{theorem}, with probability at least $1-\delta$,
\begin{align*}
 \overline{r}_n & =
 \int r_n {\rm d}\hat{\rho}_n
 + \left( \frac{\mathcal{M}_{\phi_{q},n} }{\delta}\right)^{\frac{1}{q}}
  \left(D_{\phi_p-1}(\hat{\rho}_n,\pi) +1 \right)^{\frac{1}{p}}
\\
& = \min_{\rho} \left\{
 \int r_n {\rm d}\rho
 + \left( \frac{\mathcal{M}_{\phi_{q},n} }{\delta}\right)^{\frac{1}{q}}
  \left(D_{\phi_p-1}(\rho,\pi) +1 \right)^{\frac{1}{p}} \right\}
\end{align*}
where the minimum holds for any probability distribution $\rho$ over $\Theta$.
\end{prop}

\begin{proof}[Proof of~\autoref{prop-minimizer}]
For any $\rho$ we have
\begin{align*}
\overline{r}_n - \int r_n  {\rm d}\rho
 & = \int \left[ \overline{r}_n - r_n \right] {\rm d}\rho
 \\
 & =  \int \left[ \overline{r}_n - r_n \right]_+ {\rm d}\rho
       -  \int \left[ \overline{r}_n - r_n \right]_- {\rm d}\rho
 \\
 & \leq  \int \left[ \overline{r}_n - r_n \right]_+ {\rm d}\rho
 =  \int \left[ \overline{r}_n - r_n \right]_+ \frac{{\rm d}\rho}{{\rm d}\pi}
         {\rm d}\pi
  \\
  & \leq \left(\int \left[ \overline{r}_n - r_n \right]_+^q
         {\rm d}\pi\right)^{\frac{1}{q}}
   \left( \int \left(\frac{{\rm d}\rho}{{\rm d}\pi}\right)^p
         {\rm d}\pi \right)^{\frac{1}{p}}
    \\
 & \leq \left(\frac{\mathcal{M}_{\phi_{q},n} }{\delta}\right)^{\frac{1}{q}}
        \left(D_{\phi_p-1}(\rho,\pi) +1 \right)^{\frac{1}{p}}
\end{align*}
where we used H\"older's inequality and then the definition of $\overline{r}_n$
in the last line.
Moreover, we can check that the two inequalities above become equalities when $\rho=\hat{\rho}_n$: from~\eqref{definitionRho},
\begin{align*}
\overline{r}_n - \int r_n  {\rm d}\hat{\rho}_n
 & = \int \left[ \overline{r}_n - r_n \right] {\rm d}\hat{\rho}_n
 = \int \left[ \overline{r}_n - r_n \right]_+ {\rm d}\hat{\rho}_n
 \\
& =   \frac{ \int \left[ \overline{r}_n - r_n \right]_+
\left[\overline{r}_n -r_n\right]_+^{\frac{1}{p-1}}
         {\rm d}\pi
}{
\int \left[\overline{r}_n -r_n\right]_+^{\frac{1}{p-1}} {\rm d}\pi}
 =  \frac{ \int 
\left[\overline{r}_n -r_n\right]_+^{q}
         {\rm d}\pi
}{
\int \left[\overline{r}_n -r_n\right]_+^{\frac{1}{p-1}} {\rm d}\pi}
 \\
 & = \frac{ \left(\int 
\left[\overline{r}_n -r_n\right]_+^{q}
         {\rm d}\pi \right)^{\frac{1}{p}+\frac{1}{q}}
}{
\int \left[\overline{r}_n -r_n\right]_+^{\frac{1}{p-1}} {\rm d}\pi}
 = \left(\int \left[ \overline{r}_n - r_n \right]_+^q
         {\rm d}\pi\right)^{\frac{1}{q}}
\frac{   \left( \int  \left[\overline{r}_n -r_n\right]_+^{\frac{p}{p-1}}
          {\rm d}\pi \right)^{\frac{1}{p}}}
    {\int \left[\overline{r}_n -r_n\right]_+^{\frac{1}{p-1}} {\rm d}\pi    }
\\
& = \left(\frac{\mathcal{M}_{\phi_{q},n} }{\delta}\right)^{\frac{1}{q}}
   \left( \int \left(\frac{{\rm d}\hat{\rho}_n}{{\rm d}\pi}\right)^p
         {\rm d}\pi \right)^{\frac{1}{p}}
 = \left(\frac{\mathcal{M}_{\phi_{q},n} }{\delta}\right)^{\frac{1}{q}}
        \left(D_{\phi_p-1}(\hat{\rho}_n,\pi) +1 \right)^{\frac{1}{p}}.
\end{align*}
\end{proof}

A direct consequence of~\eqref{eq-thm-1} and~\eqref{eq-thm-2} is the following
result, which provides theoretical guarantees for $\hat{\rho}_n$.

\begin{prop}
\label{proposition-toto}
Under the assumptions of~\autoref{theorem}, with probability at least $1-\delta$, 
 \begin{equation}
 \int R {\rm d}\hat{\rho}_n
   \leq
 \overline{r}_n
 \label{eq-thm-3}
  \leq \inf_{\rho} \left\{
  \int R {\rm d}\rho
 + 2 \left( \frac{\mathcal{M}_{\phi_{q},n} }{\delta}\right)^{\frac{1}{q}}
  \left(D_{\phi_p-1}(\rho,\pi) +1 \right)^{\frac{1}{p}}
  \right\}.
\end{equation}
\end{prop}

\begin{proof}[Proof of~\autoref{proposition-toto}]
 First,~\eqref{eq-thm-1} brings:
  \begin{align}
  \int R {\rm d}\hat{\rho}_n 
 & \quad \leq
 \int r_n {\rm d}\hat{\rho}_n
 + \left( \frac{\mathcal{M}_{\phi_{q},n} }{\delta}\right)^{\frac{1}{q}}
  \left(D_{\phi_p-1}(\hat{\rho}_n,\pi) +1 \right)^{\frac{1}{p}}
  \nonumber \\
 & \quad = \inf_{\rho}\left\{
 \int r_n {\rm d}\rho
 + \left( \frac{\mathcal{M}_{\phi_{q},n} }{\delta}\right)^{\frac{1}{q}}
  \left(D_{\phi_p-1}(\rho,\pi) +1 \right)^{\frac{1}{p}}
 \right\}
   \label{eq-thm-1-particulier}
\end{align}
by definition of $\hat{\rho}_n$,
and \autoref{prop-minimizer} shows that the right-hand side
is $\bar{r}_n$. Plug~\eqref{eq-thm-2} into~\eqref{eq-thm-1-particulier}
to get the desired result.
\end{proof}

\begin{exm}
 As an example of an application of~\autoref{proposition-toto}, we come back to the setting of a possibly heavy-tailed time series. More precisely, we assume that we are under the assumptions of~\autoref{coro:timeseries}. In particular, $X_i=(1,Y_{i-1})\in\mathbb{R}^2$ and $f_{\theta}(\cdot)=\left<\theta,\cdot\right>$ and $p=q=2$. For the sake of simplicity, assume that the parameter space is $\Theta = [-1,1]^2$. Let us fix $\pi$ as uniform on $[-1,1]^2$. The empirical bound stated that, with probability at least $1-\delta$, for any $\rho$,
 $$
 \int R {\rm d}\rho
 \leq
 \int r_n {\rm d}\rho
 + \sqrt{ \frac{\nu[1+\chi^2(\rho,\pi)]}{n \delta}}
 $$
 where we remind that
 \begin{align*}
 \nu & = 32 \mathbb{E}\left(Y_i^6\right)^{\frac{2}{3}} \sum_{j\in\mathbb{Z}} \alpha_j^{\frac{1}{3}} \left( 1+ 4 \int\|\theta\|^6 \pi({\rm d}\theta)\right) \\
 & \leq 1056 \mathbb{E}\left(Y_i^6\right)^{\frac{2}{3}} \sum_{j\in\mathbb{Z}} \alpha_j^{\frac{1}{3}}.
 \end{align*}
 In this context the minimizer of the right-hand side is
 \begin{equation*}
\frac{{\rm d}\hat{\rho}_n}{{\rm d}\pi}(\theta) = \frac{
\left[\overline{r}_n -r_n(\theta)\right]_+
}{
\int \left[\overline{r}_n -r_n\right]_+ {\rm d}\pi
}
\end{equation*}
where
$$
\overline{r}_n = \min\left\{u\in\mathbb{R},
\int
\left[u -r_n(\theta)\right]_+
\pi({\rm d}\theta) = \frac{\nu}{n \delta} \right\}.
$$
 The application of~\autoref{proposition-toto} leads to, with probability at least $1-\delta$,
  $$
 \int R {\rm d}\hat{\rho}_n
 \leq
 \inf_{\rho} \left\{
 \int R {\rm d}\rho
 + 2 \sqrt{ \frac{\nu[1+\chi^2(\rho,\pi)]}{n \delta}}
 \right\}.
 $$
 Note that it is possible to derive an oracle inequality from this. Let us denote by $\bar{\theta}=(\bar{\theta}_1,\bar{\theta}_2)$ the minimizer of $R$. Consider the following posteriors, for $1\leq i,j \leq N$:
 $$ \rho_{(i,j),N} \text{ is uniform on } \left[ -1 + \frac{2(i-1)}{N},-1 + \frac{2i}{N} \right] \times \left[ -1 + \frac{2(j-1)}{N},-1 + \frac{2j}{N} \right] .$$
 For $N$ fixed, there is always a pair $(i,j)$ such that $\bar{\theta}$ belongs to the support of $ \rho_{(i,j),N}$. Elementary calculus shows that, for any $\theta$ in the support of $ \rho_{(i,j),N}$ then
 $$ R(\theta) - R(\bar{\theta}) \leq \frac{2}{N} \left(1+4 \mathbb{E}(|Y_i|) + 3 \mathbb{E}(Y_i^2) \right) =: \frac{\nu'}{N} .$$
 Moreover,
 $$ 1+\chi^2(\rho_{(i,j),N},\pi) = \frac{N^2}{2} .$$
 So the bound becomes
   $$
 \int R {\rm d}\hat{\rho}_n
 \leq
 \inf_{\theta\in[-1,1]^2} R(\theta) + \inf_{N \in\mathbb{N}^*} \left\{ \frac{\nu'}{N}
 + N \sqrt{\frac{2 \nu}{ n \delta}} \right\}
 $$
 and in particular, the choice $N=\left\lceil \sqrt{\nu'\sqrt{ n \delta/(2\nu)}} \right\rceil  $ leads to
    $$
 \int R {\rm d}\hat{\rho}_n
 \leq
 \inf_{\theta\in[-1,1]^2} R(\theta) + 3 \left( \frac{2\nu \nu'^2}{n\delta} \right)^{\frac{1}{4}}
 $$
 at least for $n$ large enough to ensure $\left(\frac{n\delta}{2\nu}\right)^{\frac{1}{4}} \geq \sqrt{\nu'}$.
\end{exm}
The last example shows that it is possible in some cases to deduce from~\autoref{proposition-toto} an oracle inequality, that is, a comparison to the performance of the optimal parameter. The end of this section is devoted to a systematic derivation of such oracle inequalities, using the complexity parameter introduced in~\autoref{section-moments}, first in its empirical version, and then in its theoretical form.

\begin{thm}
\label{thm-oracle}
Under the assumptions of~\autoref{theorem} together
with~\autoref{empirical-complexity}, with probability at least $1-\delta$,
\begin{equation}
\label{upper-bound-dim}
 \int R {\rm d}\hat{\rho}_n
  \leq
 \overline{r}_n 
  \leq \inf_{\theta\in\Theta}\ \bigl\{ r_n(\theta) \bigr\} + 2 \left(\frac{\mathcal{M}_{\phi_{q},n} }{\delta} \right)^{\frac{1}{q+d}}.
\end{equation}
\end{thm}

\begin{proof}[Proof of~\autoref{thm-oracle}]
 Put
 $$\gamma = \overline{r}_n - \inf_{\theta\in\Theta}\ \bigl\{ r_n(\theta)\bigr\}. $$
 Note that $\gamma \geq 0$. Then:
\begin{equation*}
\left(\frac{\gamma}{2}\right)^{q} \pi\left\{r_n(\theta) \leq  \frac{\gamma}{2} + \inf\ r_n \right\}
 \leq
\underbrace{\int \left[\overline{r}_n -r_n\right]_+^{q} {\rm d}\pi}_{= \frac{\mathcal{M}_{\phi_{q},n} }{\delta}}
\leq
\gamma^{q} \pi\bigl\{r_n(\theta) \leq \gamma + \inf\ r_n \bigr\}.
\end{equation*}
So:
$$
\left(\frac{\gamma}{2}\right)^{q} \pi\left\{r_n(\theta)
\leq \frac{\gamma}{2} + \inf\ r_n \right\}
\leq \frac{\mathcal{M}_{\phi_{q},n} }{\delta}
$$
and, using \autoref{empirical-complexity},
$$
\left(\frac{\gamma}{2}\right)^{q} \left(\frac{\gamma}{2}\right)^d \leq \frac{\mathcal{M}_{\phi_{q},n} }{\delta}
$$
which yields:
$$
\gamma \leq 2 \left(\frac{\mathcal{M}_{\phi_{q},n} }{\delta} \right)^{\frac{1}{q+d}}.
$$
\end{proof}

We can also perform an explicit minimization of the oracle-type
bound~\eqref{eq-thm-3}, which leads to a variant of \autoref{thm-oracle}
under a non-empirical complexity assumption.

\begin{dfn}
Put
$$
\overline{R}_n = \min \left\{u \in\mathbb{R}:
\int
\left[u -R(\theta)\right]_+^{q}
\pi({\rm d}\theta) = \frac{2^q \mathcal{M}_{\phi_{q}} }{\delta} \right\}.
 $$
\end{dfn}

\begin{assum}
\label{complexity}
There exists $d>0$ such that, for any $\gamma>0$,
$$ \pi \Bigl\{ \theta\in\Theta:
R(\theta) \leq \inf_{\theta'\in\Theta}\ \bigl\{ R(\theta') \bigr\} + \gamma \Bigr\}
\geq \gamma^d .$$
\end{assum}

\begin{thm}
Under the assumptions of~\autoref{theorem} together
with~\autoref{complexity}, with probability at least $1-\delta$,
\begin{equation*}
  \int R {\rm d}\hat{\rho}_n
  \leq \overline{R}_n
  \leq \inf_{\theta\in\Theta}\ R(\theta) +
  2^{\frac{q}{q+d}} \left(\frac{\mathcal{M}_{\phi_{q},n} }{\delta} \right)^{\frac{1}{q+d}}.
\end{equation*}
\end{thm}

The proof is a direct adaptation of the proofs of~\autoref{prop-minimizer} and~\autoref{thm-oracle}.

\section{Discussion and perspectives}

We proposed a new type of PAC-Bayesian bounds, which makes use of Csisz\'ar's $f$-divergence to generalize the Kullback-Leibler divergence. This is an extension of the results in~\cite{begin2016pac}. In favourable contexts, there exists sophisticated approaches to get better bounds, as discussed in the introduction. However, the major contribution of our work is that our bounds hold in hostile situations where no PAC bounds at all were available, such as heavy-tailed time series. We plan to study the connections between our PAC-Bayesian bounds and aforementionned approaches by~\cite{mendelsonlearning2015} and~\cite{grunwald2016fast} in future works.

\subsubsection*{Acknowledgements}

We would like to thank Pascal Germain for fruitful discussions, along with two anonymous Referees and the Editor for insightful comments.

% The authors gratefully acknowledges financial support from the research programme {\it New Challenges for New Data} from LCL and GENES, hosted by the {\it Fondation du Risque}, from Labex ECODEC (ANR - 11-LABEX-0047) and from Labex CEMPI  (ANR-11-LABX-0007-01).

%\clearpage

\bibliographystyle{abbrvnat}
\bibliography{biblio}

\end{document}